\documentclass[letterpaper, 10 pt, conference]{ieeeconf}  

\IEEEoverridecommandlockouts                              

\usepackage{graphics} 
\usepackage{epsfig} 
\usepackage{amsmath} 
\usepackage{amssymb}  
\usepackage{pifont}
\usepackage{multirow}
\usepackage{xcolor}
\usepackage{tabularx}
\usepackage{comment}
\usepackage{subcaption}
\usepackage{lipsum} 
\usepackage{graphicx,subcaption}
\usepackage[font={footnotesize,it}]{caption}
\usepackage{multirow}

\usepackage{fancyhdr}
\usepackage{ragged2e}
\usepackage{hyperref}
\fancypagestyle{ieee_notice}
{
    \setlength{\headheight}{32pt}
    \addtolength{\topmargin}{-32pt}
    \lhead{\fbox{\begin{minipage}{\textwidth}
    \noindent\footnotesize\justifying\copyright 2021 IEEE DOI:\href{https://doi.org/10.1109/ICRA48506.2021.9561984}{110.1109/ICRA48506.2021.9561984}. Personal use of this material is permitted.  Permission from IEEE must be obtained for all other uses, in any current or future media, including reprinting/republishing this material for advertising or promotional purposes, creating new collective works, for resale or redistribution to servers or lists, or reuse of any copyrighted component of this work in other works.\end{minipage}}}
    \cfoot{} 
}

\makeatletter
\newcommand{\thickhline}{%
    \noalign {\ifnum 0=`}\fi \hrule height 1pt
    \futurelet \reserved@a \@xhline
}
\newcolumntype{"}{@{\hskip\tabcolsep\vrule width 1pt\hskip\tabcolsep}}
\makeatother

\title{\LARGE \bf
Learning Shape Control of Elastoplastic Deformable Linear Objects}

\author{Rita Laezza$^{1}$ and Yiannis Karayiannidis$^{1}$
\thanks{$^{1}$ Division of Systems and Control, Department of Electrical Engineering,
        Chalmers University of Technology, Sweden
        {\tt\small \{laezza , yiannis\}@chalmers.se}}%
}

\begin{document}
\maketitle
\thispagestyle{ieee_notice} 
\pagestyle{empty}

\begin{abstract}
Deformable object manipulation tasks have long been regarded as challenging robotic problems. However, until recently very little work has been done on the subject, with most robotic manipulation methods being developed for rigid objects. Deformable objects are more difficult to model and simulate, which has limited the use of model-free Reinforcement Learning (RL) strategies, due to their need for large amounts of data that can only be satisfied in simulation. This paper proposes a new shape control task for Deformable Linear Objects (DLOs). More notably, we present the first study on the effects of elastoplastic properties on this type of problem. Objects with elastoplasticity such as metal wires, are found in various applications and are challenging to manipulate due to their nonlinear behavior. We first highlight the challenges of solving such a manipulation task from an RL perspective, particularly in defining the reward. Then, based on concepts from differential geometry, we propose an intrinsic shape representation using discrete curvature and torsion. Finally, we show through an empirical study that in order to successfully solve the proposed task using Deep Deterministic Policy Gradient (DDPG), the reward needs to include intrinsic information about the shape of the DLO.
\end{abstract}

\section{Introduction}
In recent years, there has been a growing interest in deformable object grasping and manipulation problems by the robotics community \cite{Sanchez2018,khalil2010dexterous}. This is due in part to their widespread across diverse applications as well as their increased complexity, when compared to rigid object tasks. These problems have been shown to be difficult to solve with classical approaches \cite{Sanchez2018}. Consequently, learning-based methods are being explored as a more powerful alternative \cite{khalil2010dexterous}. Intuitively, if a robot is to reach human-level dexterity, there may be a need for human-inspired learning. Reinforcement Learning (RL) is a particularly promising family of methods which seeks to make robots capable of learning through experience \cite{doi:10.1177/0278364913495721}. RL has been proven successful in solving complex games, such as Go \cite{silver2016mastering}, as well as robotic control tasks, such as pick and place \cite{andrychowicz2017hindsight}.

Contrary to grasping and manipulation of rigid objects, which have been extensively addressed in the robotics literature, non-rigid objects have been largely overlooked \cite{Sanchez2018}. Though some of the same methods can be extended to particular types of deformable objects, there are still many problems left unsolved \cite{Sanchez2018}. Most notably, while manipulation of rigid objects focuses mainly on controlling their pose, when manipulating deformable objects it is often their shape which needs to be controlled \cite{khalil2010dexterous}. Furthermore, dealing with materials which are highly deformable or with elastoplastic properties, makes modeling and sensing of these objects a difficult challenge. In addition, most work on deformable object manipulation has focused on specialized tasks, from applications like robotic surgery, food processing and fabric manufacturing \cite{Sanchez2018}. While this is a practical choice to solve real-life problems, the solutions are often not general \cite{zhu2018dual}. With this work we propose a strategy for explicit shape control of elastic and/or plastic objects, which could potentially be applied to a large range of problems.

\begin{figure}[tpbb]
	\centering
	\smallskip \smallskip
	\includegraphics[width=\columnwidth]{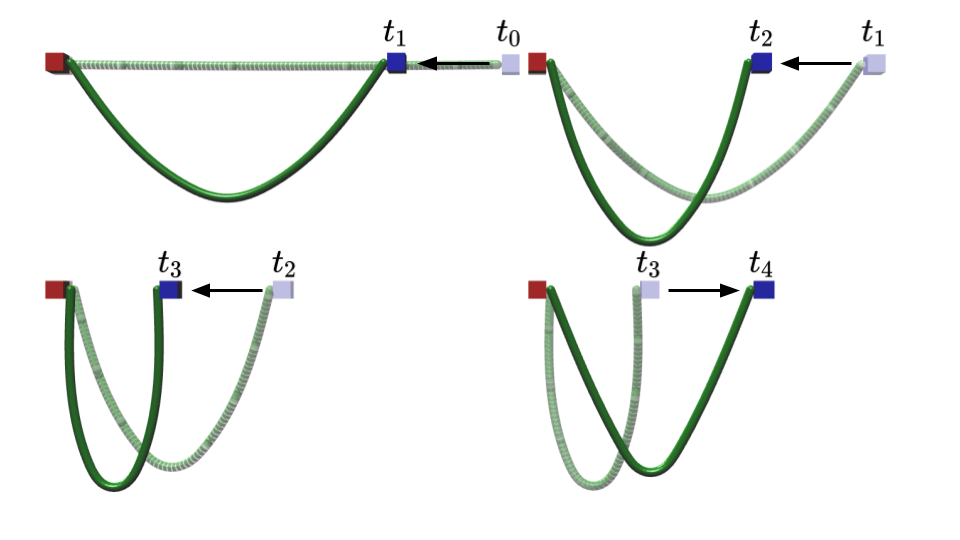}
	\caption{Simulation of DLO with plastic properties. By bending the object first inwards and then outwards, when returning the gripper to the same position at $t_2$ and $t_4$, the shape of the DLO is different due to (permanent) plastic deformation. This motion is executed along a single DoF.}
	\label{fig:plastic_deformation}
	\vspace*{-0.5cm}
\end{figure}

According to classification criteria suggested by Sanchez et al. \cite{Sanchez2018}, deformable objects can be categorized based on their mechanical properties, i.e. low or high compression strength and their geometric properties, i.e. linear, planar or solid shapes. In this work we focus on Deformable Linear Objects (DLOs) with elastoplastic properties. Objects that fit into this category include metal wires, rods and cables, found across multiple applications including medical, industrial, and household services. DLOs are an appealing choice for their relative geometric simplicity, making them more efficient to simulate but still complex to manipulate. Within this class, we found that the manipulation of objects with elastoplastic properties is yet to be studied, with most of the literature focusing on purely elastic DLOs or low compression strength DLOs such as ropes, which exhibit plastic behavior \cite{Sanchez2018}. The reason elastoplastic materials make for a particularly difficult class of objects is due to their nonlinear behavior, starting as purely elastic up to a yield point, after which transitioning to a plastic domain. This is illustrated with an example of DLO manipulation in Figure \ref{fig:plastic_deformation}, where plastic deformation occurs after the initial elastic deformation, leading to potentially irreversible changes. A practical application of elastoplastic wires can be found in the production of orthodontic braces. Our work opens up the potential for RL-based automation of the shaping process, since this is still done manually to a large extent \cite{xia2016development}. 

To address the problem of robotic manipulation of elastoplastic DLOs, we are interested on the ability to learn velocity control policies in a model-free fashion. To that end, we have implemented a simulation environment with a new shape control problem. The control policy is learned in task space and controlled by a Cartesian gripper with varying Degrees of Freedom (DoFs). The gripper grasps the object either with a fixed grasp or a flexible pinch, allowing rotation around one axis (i.e. hinge constraint). Since we aim to learn continuous actions, a policy gradient method is used, namely Deep Deterministic Policy Gradient (DDPG) \cite{lillicrap2015continuous}. 

As our main contribution, we present a detailed evaluation of the proposed shape control problem from a Reinforcement Learning perspective. We begin by presenting the challenges of designing a reward signal. We propose shape representations using concepts from discrete differential geometry, namely curvature and torsion. Based on these, we evaluate three dense reward functions in a rigorous empirical study. Further, the impact of parameters such as mechanical properties of the DLO, number of controlled DoFs and type of grasp is also studied.  


\section{Related Work} \label{related_work}
To date, ropes or rope-like objects are the most researched group of DLOs in robotic manipulation. Common problems involving ropes include knot tying, untangling, threading and reaching goal-configurations on a flat surface \cite{Sanchez2018}. While all of these present interesting challenges, only the latter represents an \textbf{explicit shape control} problem. For the others, what matters is not the final geometric deformation, but the configuration of the DLO, relative to itself, or other objects. Within these \textbf{implicit shape control} problems, the work by Berenson \cite{berenson2013manipulation}, recently extended in \cite{ruan2018accounting}, proposes promising methods which preclude the use of physical simulation.

Deformable objects simulation is still an active research topic. A great part of advances in the field come from the computer graphics community, such as Pai \cite{pai2002strands}, that used a Cosserat formulation to develop fast simulation algorithms. There have also been efforts to model DLOs with the intent to solve deformation tasks, more notably by Wakamatsu et al. \cite{wakamatsu2004static}, where a method based on differential geometry was used for motion planning. Other common DLO modeling approaches include Finite Element Methods (FEM) and Mass-Spring-Damper (MSD) systems \cite{Sanchez2018}. 

Sensing and state estimation of non-rigid objects also presents a challenge which is often tackled separately \cite{schulman2013object, javdani2011modeling}. However, recent work on robotic shape control of DLOs combines different vision-based state estimation methods with control strategies. Yan et al. \cite{yan2020self} used self-supervised learning to estimate the state of a rope resting on a tabletop, controlled by a single-arm. The manipulation was done by successive grasping and planning, after each state estimation step. Zhu et al. \cite{zhu2018dual} used Fourier series to model the DLO shape and successfully deformed flexible cables into desired shapes, using a velocity controlled dual-armed robot. There have also been different approaches for state representations of a DLO's shape, including node-graphs, Frenet coordinate frames and Kirchoff elastic rods \cite{Sanchez2018, wakamatsu2004static, bretl2014quasi}. More recently, using deep learning techniques has opened up the possibility to learn directly from the high-dimensional raw image data \cite{matas2018sim}. This can be used in end-to-end strategies, where robot joint velocities are obtained directly from pixels.

We conclude this section by highlighting works which applied RL for deformable object manipulation tasks. Clom{\'e} et al. \cite{colome2015friction} first implemented a policy improvement strategy with path integrals to manipulate a scarf around a mannequin's neck. More recently, RL was used in robot-assisted endovascular catheterization \cite{chi2018trajectory}. Both of these works employ Dynamic Movement Primitives (DMPs) and Learning from Demonstration (LfD). Matas et al. \cite{matas2018sim} produced promising results in cloth-manipulation using a state-of-the-art RL algorithm. Their work was formulated both in an end-to-end manner and for sim-to-real transfer. They used a variation of Deep Deterministic Policy Gradient, named DDPG from Demonstrations (DDPGfD) which seeds the learning with expert demonstrations.  Conversely, Wu et al. \cite{wu2019learning} proposed to solve pick-and-place tasks of deformable objects completely from scratch. In contrast with our work, \cite{matas2018sim}, \cite{colome2015friction} and \cite{chi2018trajectory} covered implicit shape control problems, while \cite{wu2019learning} did not address permanent deformation.

\section{Background} \label{background}
Before presenting the proposed shape control problem, we introduce the necessary technical background on RL in \ref{reinforcement_learning} and DLO simulation in \ref{simulation}. 

\subsection{Reinforcement Learning} \label{reinforcement_learning}
In RL, control problems are framed as Markov Decision Processes (MDPs). We consider an infinite-horizon discounted MDP, defined as a tuple $(\mathcal{S},\mathcal{A},p,r, \gamma)$, where $\gamma$ is the discount factor and $\mathcal{S}$ and $\mathcal{A}$ are continuous state and action spaces, respectively. For most real-life problems this MDP is unknown since the probability density function $p(s_{t+1} | s_t, a_t)$, depends on an environment which is difficult to model. This function represents the probability of transitioning to state $s_{t+1}$, given the current state $s_t$ and action $a_t$, with $s_t, s_{t+1} \in \mathcal{S}$ and $a_t \in \mathcal{A}$. Further, in practical applications, the reward function $r: \mathcal{S} \times \mathcal{A} \rightarrow \mathbb{R}$, is defined based on the desired task, taking the environment into consideration. The reward at each transition, $r_t$, is assumed to be a bounded scalar \cite{haarnoja2018soft}. To provide a measure of success, the return at time $t$ is defined as the sum of discounted future rewards:
\begin{equation}
	G_t = \sum_{k=t}^{\infty} \gamma^{k-t} r (s_k,a_k)
\end{equation}

In policy gradient methods the objective is to find an optimal stochastic policy, $\pi_\vartheta: \mathcal{S} \rightarrow \mathcal{P}(\mathcal{A})$, which maps states to action probabilities. The optimal parameterized policy maximizes the expected return, i.e. $J(\pi) = \mathbb{E} [G_0 | \pi] $. For a specific state $s_t$ and action $a_t$, the expected return is defined as the action-value function $Q^\pi$:
\begin{equation}
	Q^\pi (s_t, a_t) = \mathbb{E}_{r_k, s_k \sim \rho^\pi, a_k \sim \pi } [G_t |  s_t, a_t]
\end{equation}
with $k \geq t$ and $\rho^\pi$ the state distribution under policy $\pi$. 

As the name indicates, DDPG learns a deterministic policy $\mu_\vartheta: \mathcal{S} \rightarrow \mathcal{A}$, instead of a stochastic one. This algorithm is considered an actor-critic method, because the policy (actor) parameters are updated based on an estimated value function (critic) \cite{lillicrap2015continuous}. Both actor and critic are modeled as Deep Neural Networks (DNNs). Parameters $\vartheta \in \mathbb{R}^n$, are updated via stochastic gradient ascent, to maximize the $Q_\varphi$ value, with $\varphi \in \mathbb{R}^m$. The policy update is calculated based on the current estimate of the $Q$ value:
\begin{equation} \label{policy_gradient}
    \nabla_\vartheta J (\mu_\vartheta) \approx \mathbb{E}_{s_t \sim \rho^\mu} [\nabla_\vartheta \mu_\vartheta (s) \nabla_a Q^\mu(s,a) |_{s=s_t,a=\mu_\vartheta(s_t)} ]
\end{equation}
Parameters of the $Q_\varphi$ network are updated according to the Bellman equation, by minimizing the loss $\mathcal{L}(\varphi)$: 
\begin{align} \label{bellman}
    \mathcal{L}(\varphi) &= \mathbb{E}_{s_{t} \sim \rho^\mu, a_t \sim \mu, r_t} \left[ (Q_\varphi(s_{t}, a_{t}) - y_t )^2 \right]  \\
    y_t &= r(s_t,a_t) + \gamma Q_\varphi(s_{t+1},\mu(s_{t+1})) \nonumber
\end{align}

To ensure sufficient exploration while experience is being sampled, noise is added to the actor policy $\beta(s_t) = \mu_\vartheta (s_t) + \mathcal{N}$, effectively making this an off-policy method. Practically, this means that the states in equations \eqref{policy_gradient} and \eqref{bellman} are sampled from $\rho^\beta$ instead of $\rho^\mu$. Moreover, experience sampled by following the exploration policy $\beta$ is stored in a replay buffer, as tuples $(s_t,a_t,r_t,s_{t+1})$. Actor and critic networks are updated by uniformly sampling mini-batches from the replay buffer, which helps mitigate problems such as learning from temporally correlated data (environment steps are not i.i.d.) and catastrophic forgetting \cite{lillicrap2015continuous}. The replay ratio defines the number of gradient updates per environment step (i.e. how much experience is trained on before being discarded). To increase stability, two sets of networks are kept so that actor and critic updates are done with respect to slow-changing target networks, with parameters $\vartheta'$ and $\varphi'$. To that end, soft updates $\varphi' \leftarrow \lambda \varphi + (1-\lambda)\varphi'$ and $\vartheta' \leftarrow \lambda \vartheta + (1-\lambda)\vartheta'$ are used for each parameterized function, with $\lambda << 1$.

\subsection{DLO Simulation} \label{simulation}
Although our aim is to implement robot learning in real-life experiments, it can be intractable to train RL algorithms directly in a real robot, since they require a lot of sampled experience. This is especially true when learning from scratch and using model-free methods, such as DDPG, which are notoriously sample inefficient \cite{haarnoja2018soft}. It is therefore common-practice to tackle problems first in simulation, and later apply sim-to-real transfer. With this in mind, we have implemented a virtual environment, to evaluate the potential of these methods for deformable object manipulation. 

When choosing a physics engine, there are many factors to consider, such as accuracy, speed and development time. One requirement added by our particular application is the need for deformable object simulation capabilities. The robotics and classical control environments available in Gym \cite{1606.01540} were implemented using MuJoCo (Multi-Joint dynamics with Contact) \cite{Todorov2012MuJoCoAP}, which seems to be the predominant choice in the Reinforcement Learning community. It offers support for three types of soft bodies, namely rope, cloth and sponge-like 3D objects. Bullet is the preferred open source alternative, which is supported both by Gazebo and V-REP \cite{v-rep}. It provides limited functionalities, although it was used for cloth simulation in \cite{matas2018sim}. SOFA (Simulation Open Framework Architecture) \cite{sofa} on the other hand is a framework targeted at interactive computational medical simulation, with good support for soft tissues. Nevertheless, we found that for DLO simulations, AGX Dynamics provides the best set of tools. 

\begin{figure}[tp]
	\vspace*{0.2cm}
	\centering
	\includegraphics[width=0.8\columnwidth]{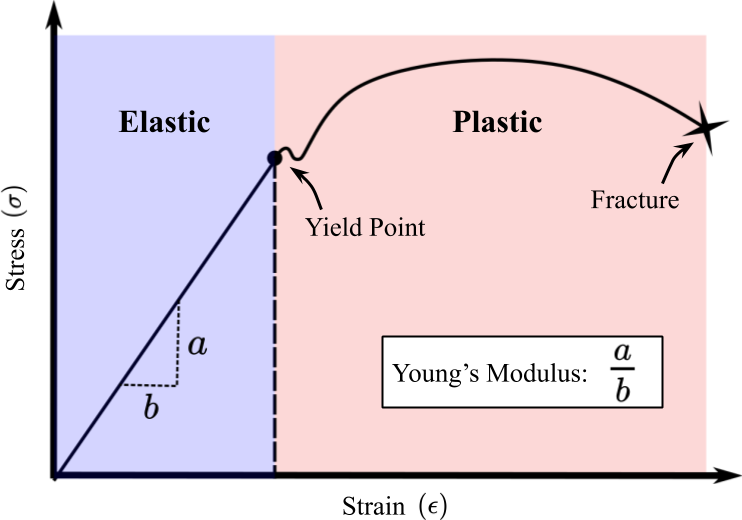}
	\caption{Illustration of typical stress-strain curve of material with elastic and plastic properties. Blue region consists of the (linear) elastic domain, while the red region consists of the (nonlinear) plastic domain.}
	\label{fig:stress_strain}
	\vspace*{-0.5cm}
\end{figure}

AGX Dynamics also offers real-time rendering and a \texttt{Cable} class which consists of a lumped element model with support for elastoplastic deformations \cite{servin2008rigid,servin2006interactive}. Further, it provides the possibility to define the object's \texttt{Material} with physically motivated mechanical properties such as Poisson's ratio, Young's modulus and the yield point, where there is the transition between elastic and plastic deformation, as illustrated in Figure \ref{fig:stress_strain}. On the other hand, purely elastic DLOs exhibit linear behavior which makes robotic shape control tasks significantly simpler.

\section{Problem Statement} \label{problem_statement}

We propose a shape control problem of an elastoplastic DLO, with two grippers holding it in free space (without obstacles), as shown in Figure \ref{fig:plastic_deformation}. The goal is to deform the DLO into a desired shape from an undeformed starting state. For simplicity, one of the grippers is static (red) while the other is able to move (blue). The control input is the linear velocity of a Cartesian gripper along each controlled DoF. This can be seen as task space velocity-resolved control of a robotic arm. We consider different number of controlled DoFs, which affects the size of both the state and action spaces. A perfect grip without translational slippage is assumed, with two modalities: hinge or lock. The former passively allows rotation about one axis of the gripping point, while the latter is completely fixed, leading to more pronounced deformations. 

In order to successfully apply RL to any application, a reward signal must be designed such that it encodes the actual goal, without inadvertently leading to high rewards in non-goal states \cite{sutton2018reinforcement}. For the proposed shape control problem, the goal can be described as a perfect overlap between the state of the achieved and the desired DLO. This is a challenging task compared to rigid body problems, where the state of an object can be summarized by its pose in $\mathbb{R}^6$. On the other hand, a perfect match for a DLO requires a state representation at least in $\mathbb{R}^{3N}$, where $N$ is the number of discrete points used to describe the DLO's shape as a point cloud. If we consider a sparse reward where a positive scalar is attributed only when the goal is reached, this can be problematic due to two main reasons:

i. When is the shape reached? This requires some distance measure which is intimately related with the DLO's shape representation. The simplest choice is to take the Euclidean distance between discrete points of the desired and achieved deformations. However, this also requires a choice of success threshold, which affects both the learning process and the accuracy of the achieved shape. 

ii. With an increased state space, exploration becomes more challenging, particularly with sparse rewards. Indeed, this may result in what is called the \textit{plateau problem}, in which the agent never experiences a positive reward during training, leading to failure to learn \cite{sutton2018reinforcement}. Note that the more complex the shape, or the greater accuracy is desired, the larger $N \gg 6$ must be, resulting in a larger state space.

\begin{figure}[tb]
    \vspace*{0.2cm}
    \begin{subfigure}{.49\columnwidth}
    \centering
    \includegraphics[width=\textwidth]{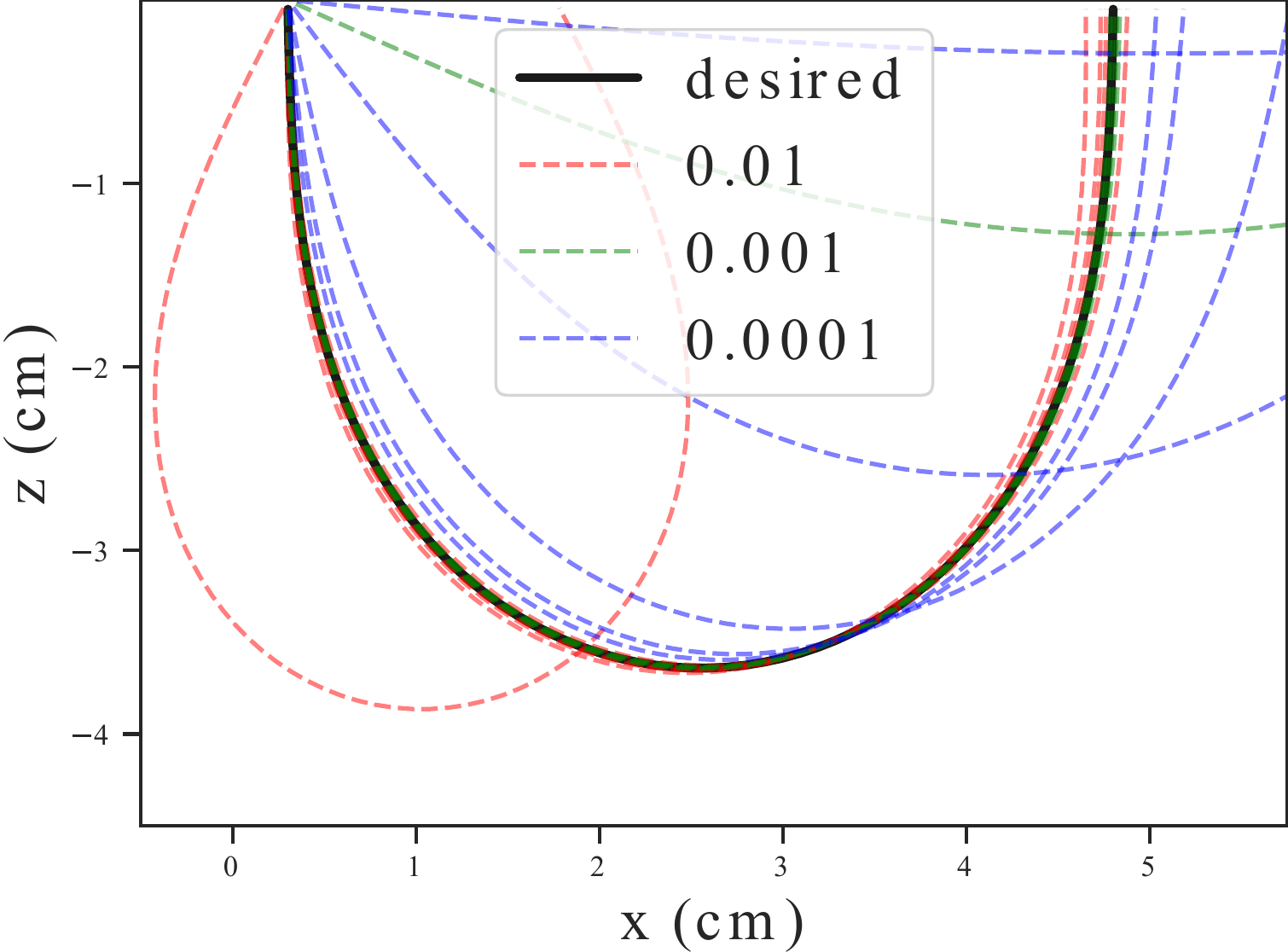}
    \end{subfigure}
    \begin{subfigure}{.49\columnwidth}
    \centering
    \includegraphics[width=\textwidth]{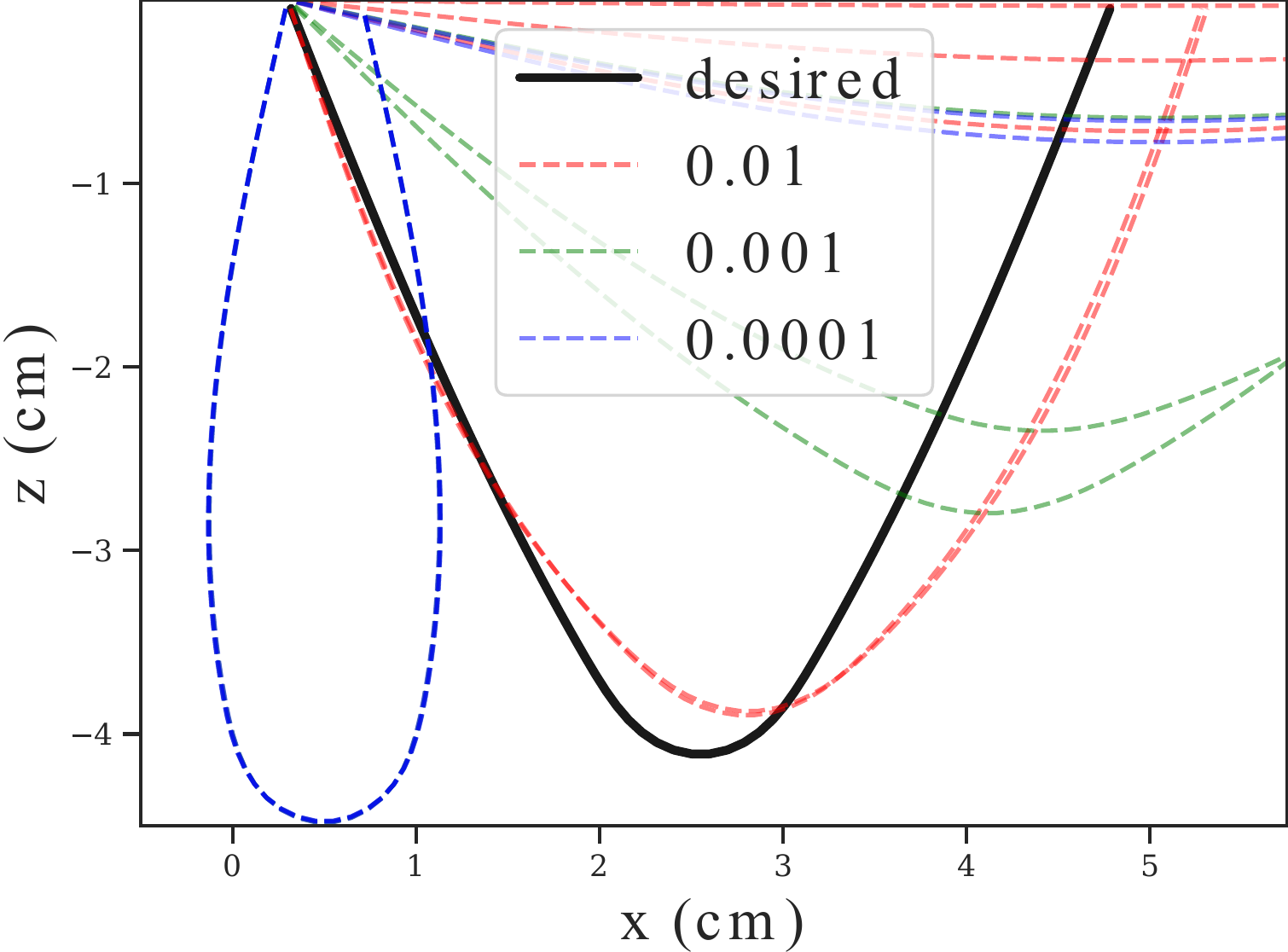}
    \end{subfigure}
    \begin{subfigure}{.49\columnwidth}
    \centering
    \includegraphics[width=\textwidth]{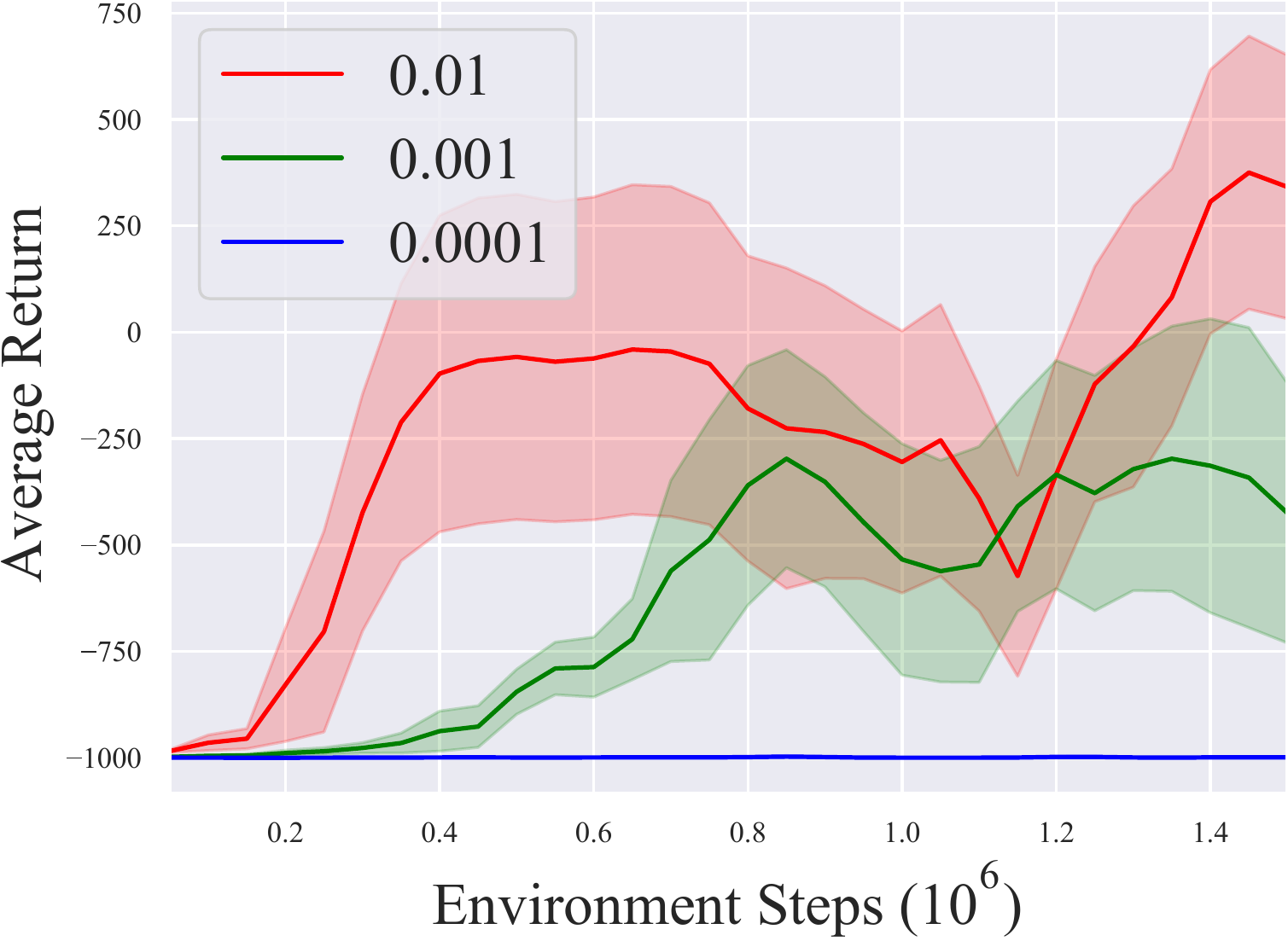}
    \caption{Elastic DLO}
    \end{subfigure}
    \begin{subfigure}{.49\columnwidth}
    \centering
    \includegraphics[width=\textwidth]{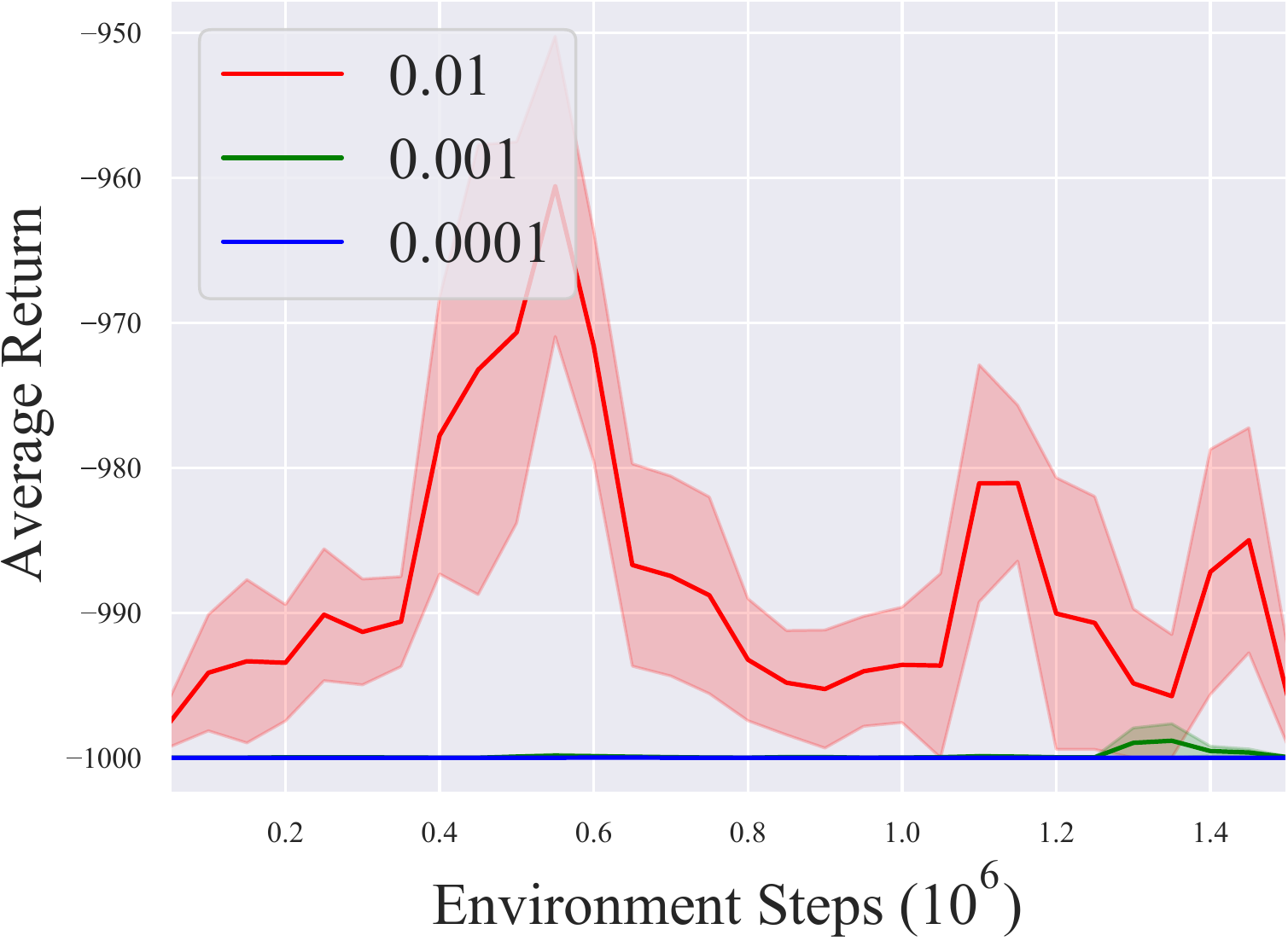}
    \caption{Elastoplastic DLO}
    \end{subfigure}
    \caption{Learning was possible for the purely elastic DLO but failed for the elastoplastic case, as shown on the top row with the final shapes obtained for 5 trials with different success thresholds. Bottom row shows the average return during training, highlighting the effect of threshold selection, with complete failure of learning for higher accuracies. Three threshold values were tested and shaded area shows standard deviation.}
    \label{fig:sparse}
    \vspace*{-0.5cm}
\end{figure}

To demonstrate these challenges we consider the proposed shape control problem, with 1 DoF control. A DDPG agent is trained with a reward of $1$ attributed only when the Euclidean distance between the desired and the achieved shape is within a given threshold; otherwise, the agent receives a reward of $-1$ at every step, encouraging the agent to reach the goal as fast as possible. Positive rewards are therefore sparse. Figure \ref{fig:sparse} (a) shows the results for an elastic DLO where the $0.001$ threshold leads to good results but the larger threshold leads to inaccurate shapes and the lower one hinders learning. Figure \ref{fig:sparse} (b) shows that for the elastoplastic DLO, even with the highest threshold i.e. lowest accuracy, the agent was unable to learn, with all trials leading to wrong shapes.


\section{Shape Representation} \label{state_representation}

Sensing of deformable objects is a challenging research topic. In Section \ref{related_work}, we list some of the state estimation methods that have been used to track the shape of DLOs. However, here we do not focus on estimation, but rather on the choice of shape representation. Given that we work in simulation, the state of the DLO in Euclidean space can be easily summarized as a point cloud with the coordinates of the lumped elements making up the object. We leave the task of extracting this point cloud from vision data as future work. From this simple representation we present useful concepts from differential geometry that can better describe the intrinsic shape of a DLO.

\begin{figure}[thpb]
	\centering
	\includegraphics[width=\columnwidth]{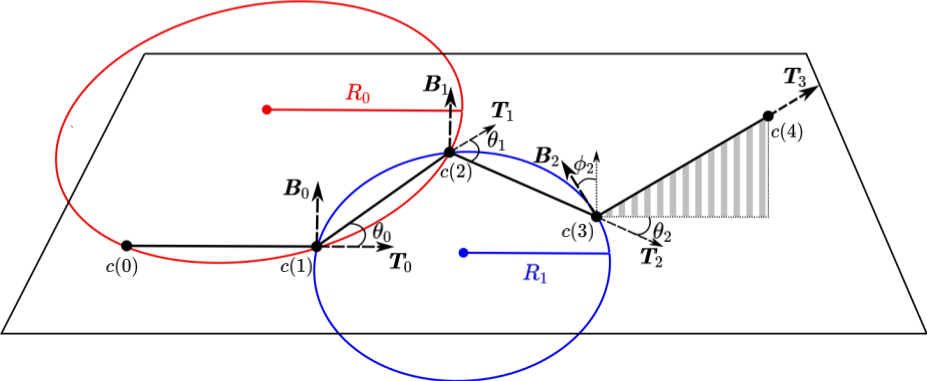}
	\caption{Discrete curve described by points $c(i)$, tangent vectors $\pmb{T}_{i}$, binormal vectors $\pmb{B}_{i}$ and angles $\theta_i$, $\phi_i$. Two osculating circles are illustrated to show the inverse relationship between radius $R_i$ and curvature $\kappa_i$.}
	\label{fig:osculating_circle}
\end{figure}

If we consider the point could of a DLO to be a discrete curve $c: \mathbb{N} \rightarrow \mathbb{R}^3$ with $N \geq 4$ points, it is possible to find a shape representation based on the notions of curvature and torsion. As shown in Figure \ref{fig:osculating_circle}, the discrete curvature $\kappa_i$ can be described through the circumscribed osculating circle. For three consecutive (noncollinear) points, there is a unique circle circumscribing them, with radius $R_i > 0$. Curvature is defined as $\kappa_i = 1 / R_i$, and for discrete curves it can be approximated as:
\begin{equation}
	\label{eq:curvature}
	\kappa_i = \frac{2}{l}\tan \frac{\theta_i}{2} \approx \frac{\theta_i}{l}, \ \ \text{with} \ \theta_i \in \left[-\frac{\pi}{2}, \frac{\pi}{2}\right]
\end{equation}
where $l$ is the segment length, and $\theta_i$ is the angle between tangent vectors of two consecutive segments. For collinear points the curvature is zero. Further the discrete torsion can be approximated as,
\begin{equation}
	\label{eq:torsion}
	\tau_i =  \frac{2}{l}\tan \frac{\phi_i}{2} \approx \frac{\phi_i}{l}, \ \ \text{with} \ \phi_i \in \left[-\frac{\pi}{2}, \frac{\pi}{2}\right]
\end{equation}
where $\phi_i$ is the angle between two consecutive binormal vectors. It is assumed that all segments have equal length \cite{carroll2013survey}. To obtain the exterior angles, for each pair of adjacent points the tangent vector needs to be computed:
\begin{equation}
    \pmb{T}_i= \frac{c(i+1) - c(i)}{l}, \ \ \ i=0, \ldots, N-1
\end{equation}
Then, for each pair of consecutive tangent vectors, the curvature angle can be obtained, 
\begin{equation}
	\theta_i = \arccos \left( \frac{ \pmb{T}_i \times  \pmb{T}_{i+1}}{|| \pmb{T}_i|| \cdot || \pmb{T}_{i+1}||} \right), \ \ \ i=0, \ldots, N-2
\end{equation}
which is enough to approximate the discrete curvature $\kappa_i$, as in equation \eqref{eq:curvature}. For the torsion, it is further necessary to compute the binormal vectors, which are orthogonal to the plane defined by the tangent and normal vectors. This can also be computed based on the plane characterized by two consecutive tangent vectors:
\begin{equation}
	\pmb{B}_{i} = \pmb{T}_i \times \pmb{T}_{i+1}, \ \ \ i=0, \ldots, N-2
\end{equation}
Finally, the torsion angle can be computed as,
\begin{equation}
	\phi_i = \arccos \left( \frac{\pmb{B}_i \times \pmb{B}_{i+1}}{||\pmb{B}_i|| \cdot ||\pmb{B}_{i+1}||} \right) , \ \ \ i=0, \ldots, N-3
\end{equation}

Based on the definitions presented in this section we move on to Section \ref{rl_formulation} where we formulate different MDP state and reward definitions which can be used to solve the proposed shape control problem.  

\section{RL Formulation} \label{rl_formulation}

We define the action space $\mathcal{A}$, such that $a \in [-1,1]$ and its dimensionality depends on the number of controlled degrees of freedom. Outputs from the policy DNN are then rescaled into viable velocity commands. The state $s$ includes the position and velocity of the end-effector, denoted by $\pmb{p}_{ee},\pmb{v}_{ee} \in \mathbb{R}^3$ respectively. Further, depending on the reward definition $r(s)$, the achieved curve $\pmb{c} = [c(i), \ldots ,c(N)]$, the achieved curvature $\pmb{\kappa} = [\kappa_i, \ldots, \kappa_{N-2}]$ and/or torsion $\pmb{\tau} = [\tau_i, \ldots, \tau_{N-4}]$ may also be included in the state definition. 

As mentioned in Section \ref{problem_statement}, the simplest distance measure is the Euclidean distance $L(\pmb{c})$ between the achieved curve $\pmb{c}$ and the desired curve  $\bar{\pmb{c}}$. This leads to our first reward function:
\begin{equation}\label{eq:position_reward}
    r_t (\pmb{c}_t) = -  L_t (\pmb{c}_t) = - || \pmb{c}_t - \bar{\pmb{c}} ||_2
\end{equation}
For planar deformations, an alternative reward function can be defined based on the desired curvature $\bar{\pmb{\kappa}}$:
\begin{equation}\label{eq:curvature_reward}
    r_t (\pmb{\kappa}_t) = - L_t (\pmb{\kappa}_t) = - || \pmb{\kappa}_t - \bar{\pmb{\kappa}} ||_2
\end{equation}
This reward can also be extended with the distance between the achieved $\pmb{\tau}$ and desired $\bar{\pmb{\tau}}$ torsion, for 3D deformations. Finally, we formulate a weighted reward function which combines the rewards from equations \eqref{eq:position_reward} and \eqref{eq:curvature_reward}:
\begin{equation}\label{eq:position_curvature_reward}
    r_t (\pmb{c}_t, \pmb{\kappa}_t) = - (1-\alpha) L_t (\pmb{c}_t) - \alpha L_t (\pmb{\kappa}_t)
\end{equation}
with $\alpha \in [0,1]$. We denote variables dependent on the current environment step with subscript $t$.

In Section \ref{experiments} we evaluate the performance of the proposed dense reward functions. To that end, we concatenate the shape representations used in the reward i.e. $\pmb{c}_t$, $\pmb{\kappa}_t$ and/or $\pmb{\tau}_t$, with $\pmb{p}_{ee}$ and $\pmb{v}_{ee}$ in a one-dimensional state vector. 

\section{Experimental Results} \label{experiments} 

\begin{figure}[tb]
    \begin{subfigure}{.49\columnwidth}
    \centering
    \includegraphics[width=\textwidth]{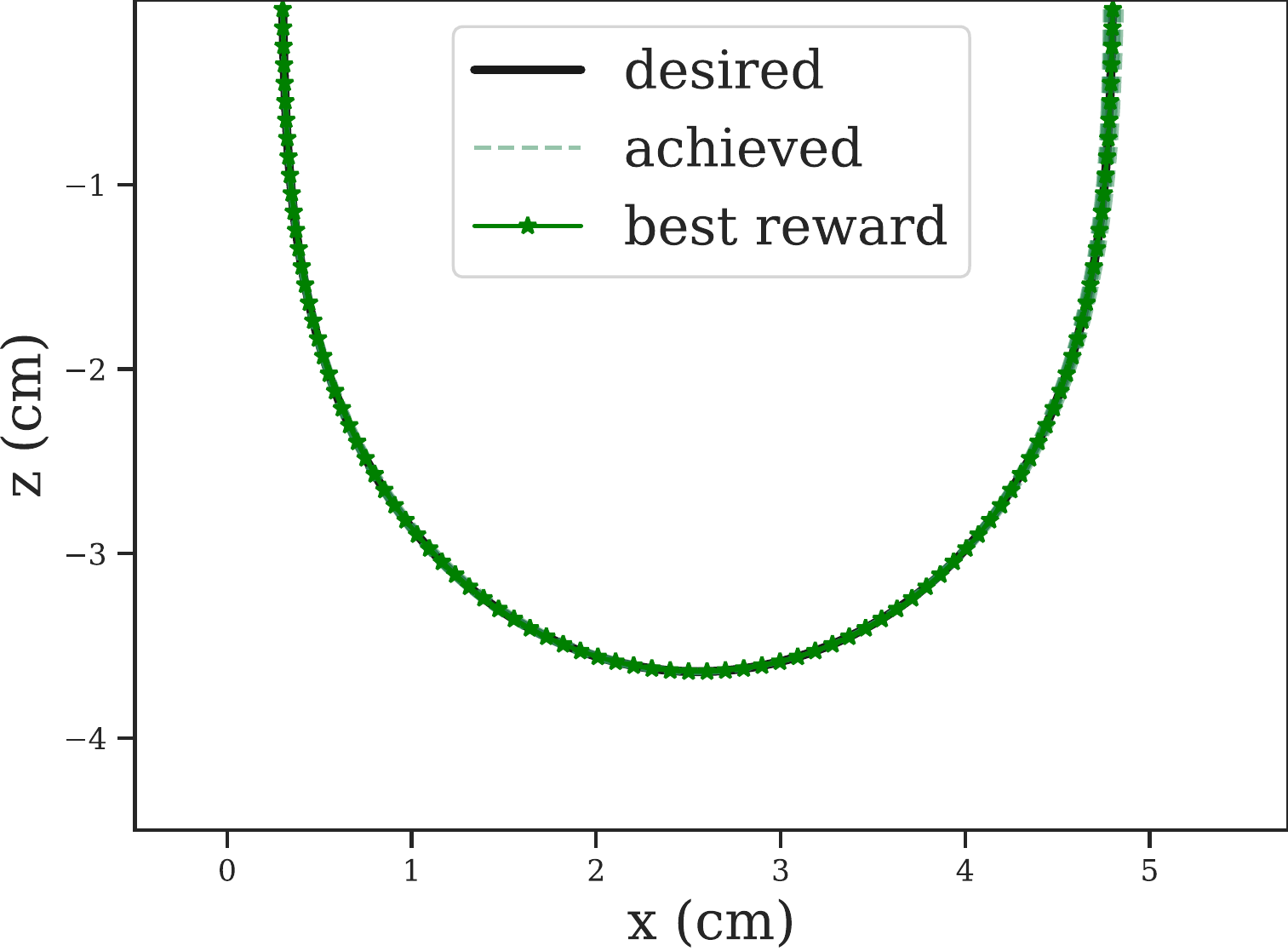}
    \caption{Hinge}
    \end{subfigure}
    \begin{subfigure}{.49\columnwidth}
    \centering
    \includegraphics[height=3.5cm]{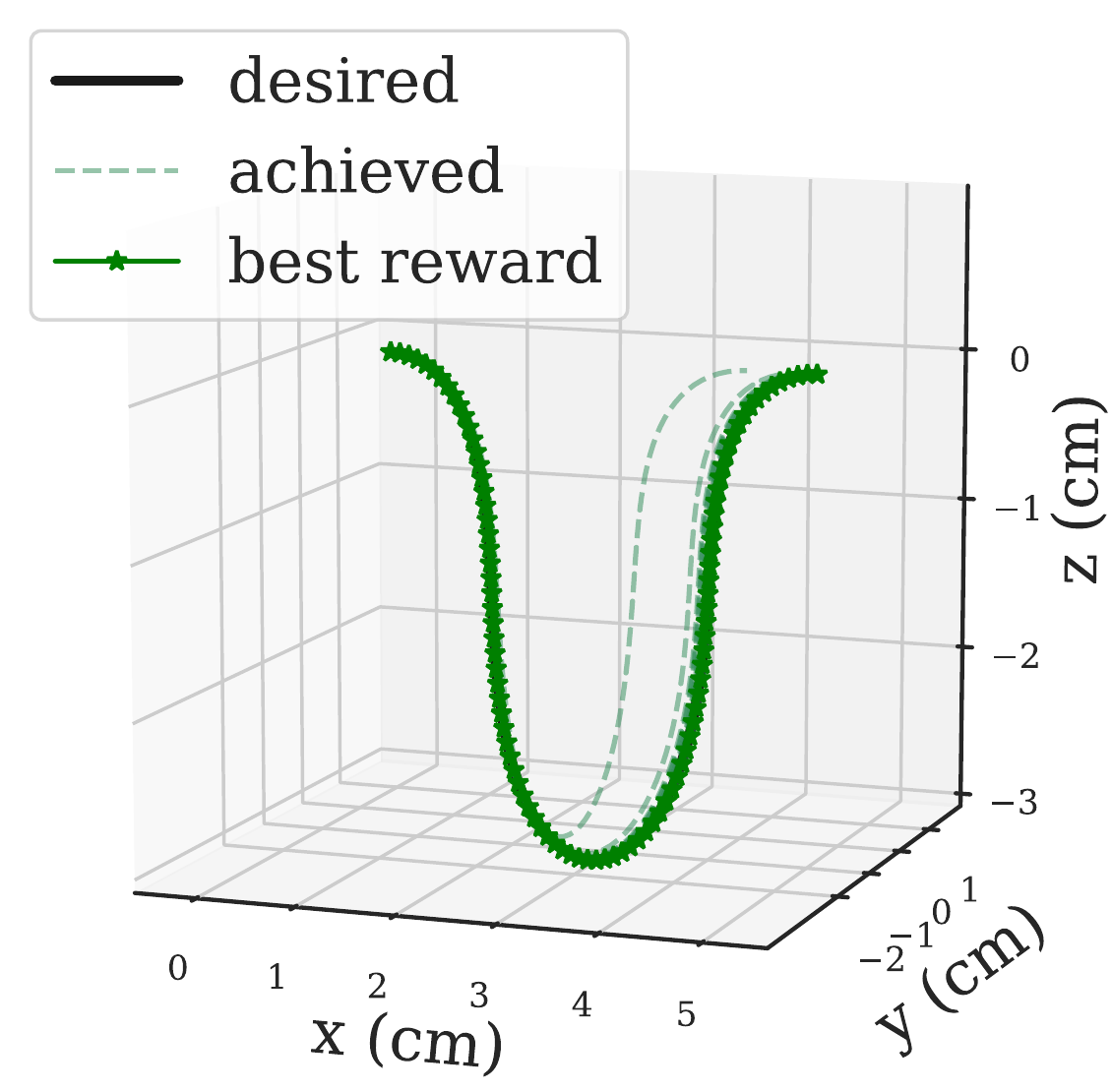}
    \caption{Lock}
    \end{subfigure}
    \caption{Results of ten trained policies to reach a desired shape with a purely elastic DLO. The same 1 DoF trajectory was implemented to generate both goals, although in (a) a hinge grip was used allowing for a planar deformation while in (b) the grip was fixed, leading to a 3D deformation.}
    \label{fig:elastic}
    \vspace*{-0.5cm}
\end{figure}

To generate the goal shapes, a 5th order polynomial trajectory is used such that plastic deformation occurs, when simulating an elastoplastic DLO. The same trajectory leads to different shapes depending on the type of grip and material properties of the object. The gripping points on the DLO are simulated as being attached on each extremity to a rigid object, by either a \texttt{Hinge} or a \texttt{Lock} constraint. We further use three \texttt{Prismatic} constraints, to simulate a Cartesian gripper. The DLO is modeled as an Aluminum cable, which is $10 \mathrm{cm}$ long and has a radius of $1 \mathrm{mm}$. The resolution of \texttt{Cable} is set to 1000 segments per meter, while its Young's Modulus is set to $69 \mathrm{MPa}$ and its Poisson's ratio to $0.35$. For the elastoplastic behavior, a yield point of $50 \mathrm{MPa}$ is defined in the bend direction of the DLO. Furthermore, the gripper is constrained by limiting its force range. The simulation time-step is set to $0.01 \mathrm{s}$, while actions are applied every second step leading to a control frequency of 50 Hz. 


In this work we evaluate the proposed shape control problem and the challenges of reward signal design using DDPG. For our experiments, the open source rlpyt \cite{stooke2019rlpyt} codebase is used. The actor and critic DNNs have the same architecture, namely two hidden layers with $[400, 200]$ neurons. The Adam optimization algorithm is used for gradient updates with learning rates of $1\times10^{-4}$ and $1\times10^{-3}$ for actor and critic, respectively. A batch size is set to 1024, sampled uniformly from the replay buffer with $5\times10^{5}$ tuples, and replay ratio of 64. Other important hyperparameters include the discount factor, $\gamma = 0.99$ and soft update rate, $\lambda=0.01$.

For each algorithm, 10 trials are performed and the results averaged. For each trial, 20 parallel agents were used to gather experience, each with a different random seed. We present the final shapes obtained for each trial, trained with 1 million environment samples. Note however that this was fixed from the start and in simpler tasks the algorithm converged earlier. Conversely, for the higher dimensional tasks training may have been insufficient. 

To highlight the challenge of elastoplasticity, we first evaluate the performance of DDPG for the same shape control problem applied to a purely elastic DLO. Results are shown in Figure \ref{fig:elastic}. In this case the agent is able to easily learn the control policy, using reward \eqref{eq:position_reward}.
\bgroup
\def\arraystretch{1.3}
\begin{table}[h!]
\centering
\caption{Results for each reward proposed in \ref{rl_formulation}, showing the final distance, $ L_T (\pmb{c}_T)$. Mean, standard deviation and best results are listed.}
\begin{tabular}{c|cc|cc}
\thickhline
\multirow{2}{*}{$r_t$} & \multicolumn{2}{c|}{\textbf{Hinge}} & \multicolumn{2}{c}{\textbf{Lock}} \\
                  & mean $\pm \ \sigma$ & best & mean $\pm \ \sigma$ & best \\ \hline
\eqref{eq:position_reward} &  $0.0132 \pm 0.0022$ & $0.0100$ & $0.0077 \pm 0.0003$ & $0.0073$ \\
\eqref{eq:curvature_reward}  & $0.0561 \pm 0.0298$ & $0.0087$ & $0.0298 \pm 0.0231$ & $\textbf{0.0010}$ \\
\eqref{eq:position_curvature_reward}  & $0.0143 \pm 0.0106$ & $\textbf{0.0019}$ & $0.0077 \pm 0.0043$ & $0.0026$ \\
\thickhline
\end{tabular}
\label{tab:results}
\end{table}
\bgroup

\begin{figure}[tb]
    \vspace*{0.2cm}
    \begin{subfigure}{.49\columnwidth}
    \centering
    \includegraphics[width=\textwidth]{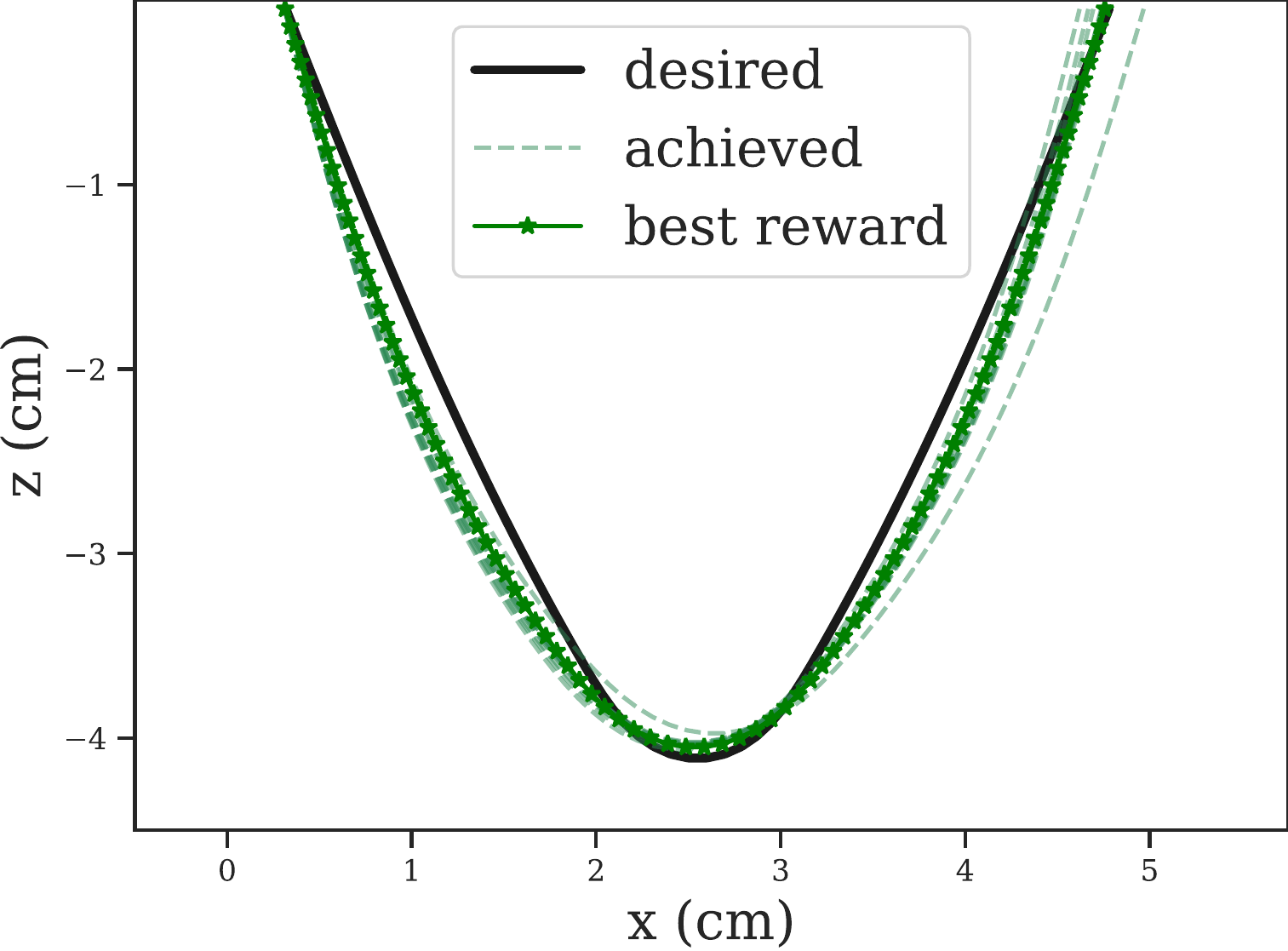}
    \end{subfigure}
    \begin{subfigure}{.49\columnwidth}
    \centering
    \includegraphics[width=\textwidth]{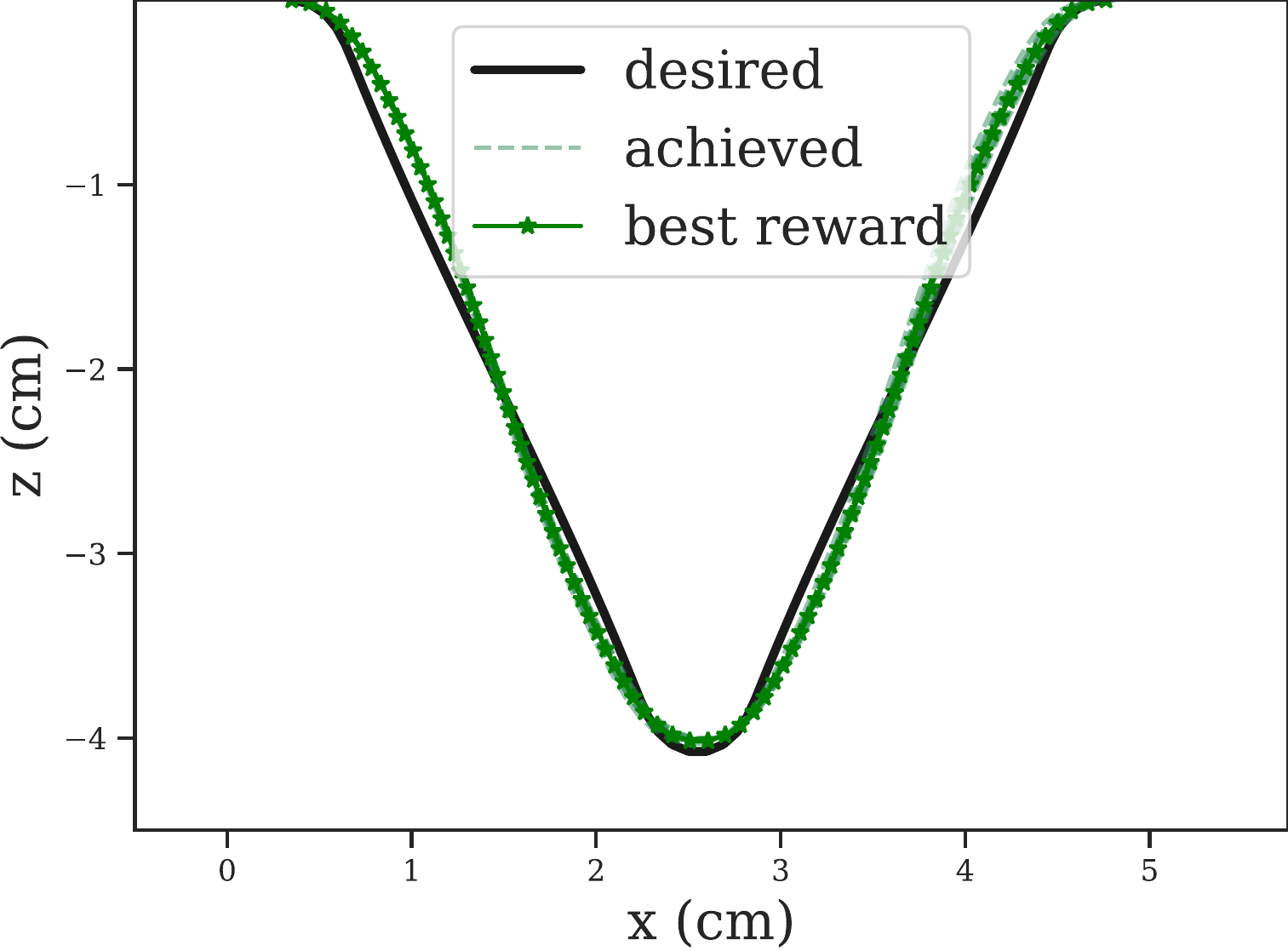}
    \end{subfigure}
    \begin{subfigure}{.49\columnwidth}
    \centering
    \includegraphics[width=\textwidth]{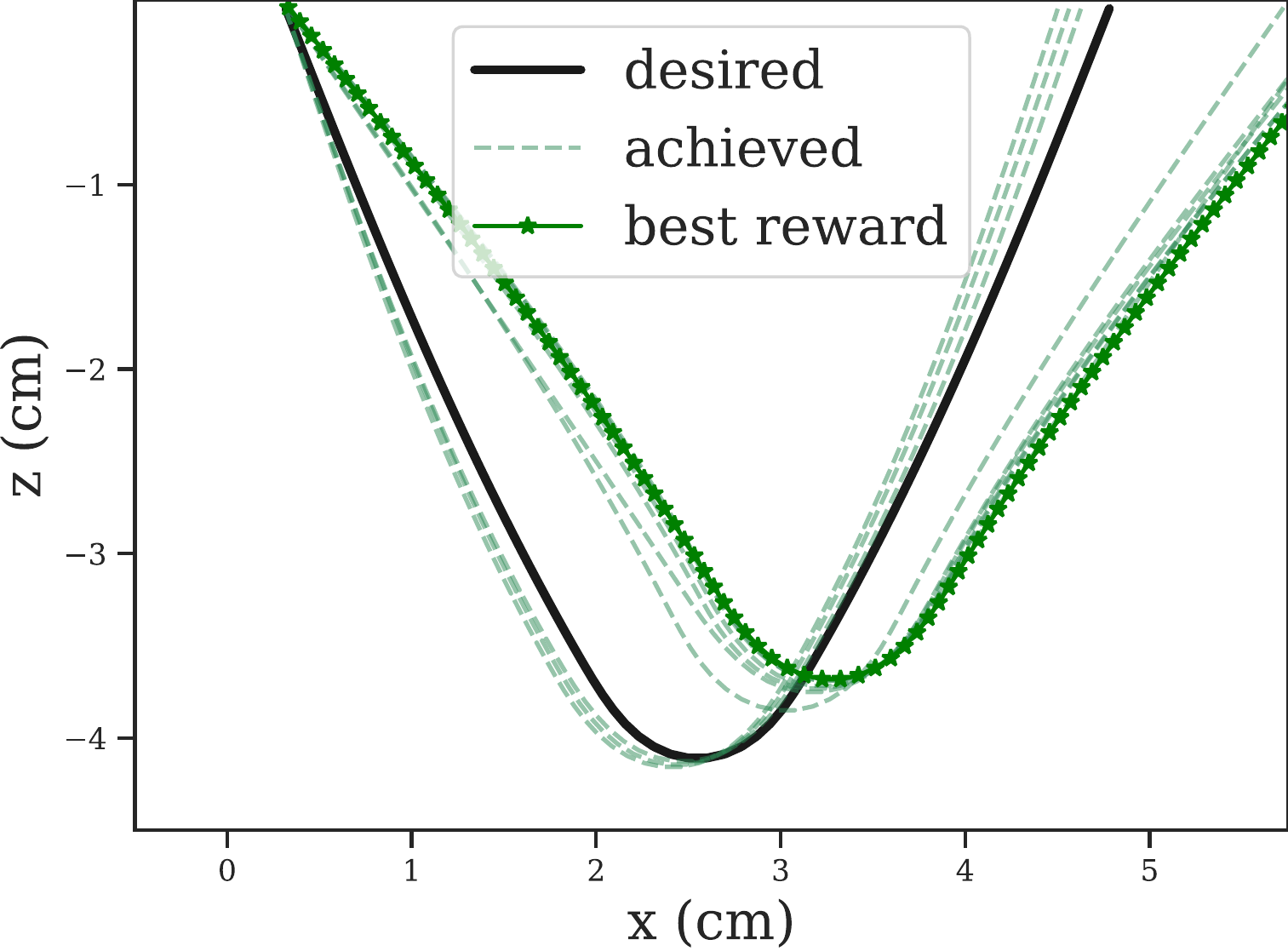}
    \end{subfigure}
    \begin{subfigure}{.49\columnwidth}
    \centering
    \includegraphics[width=\textwidth]{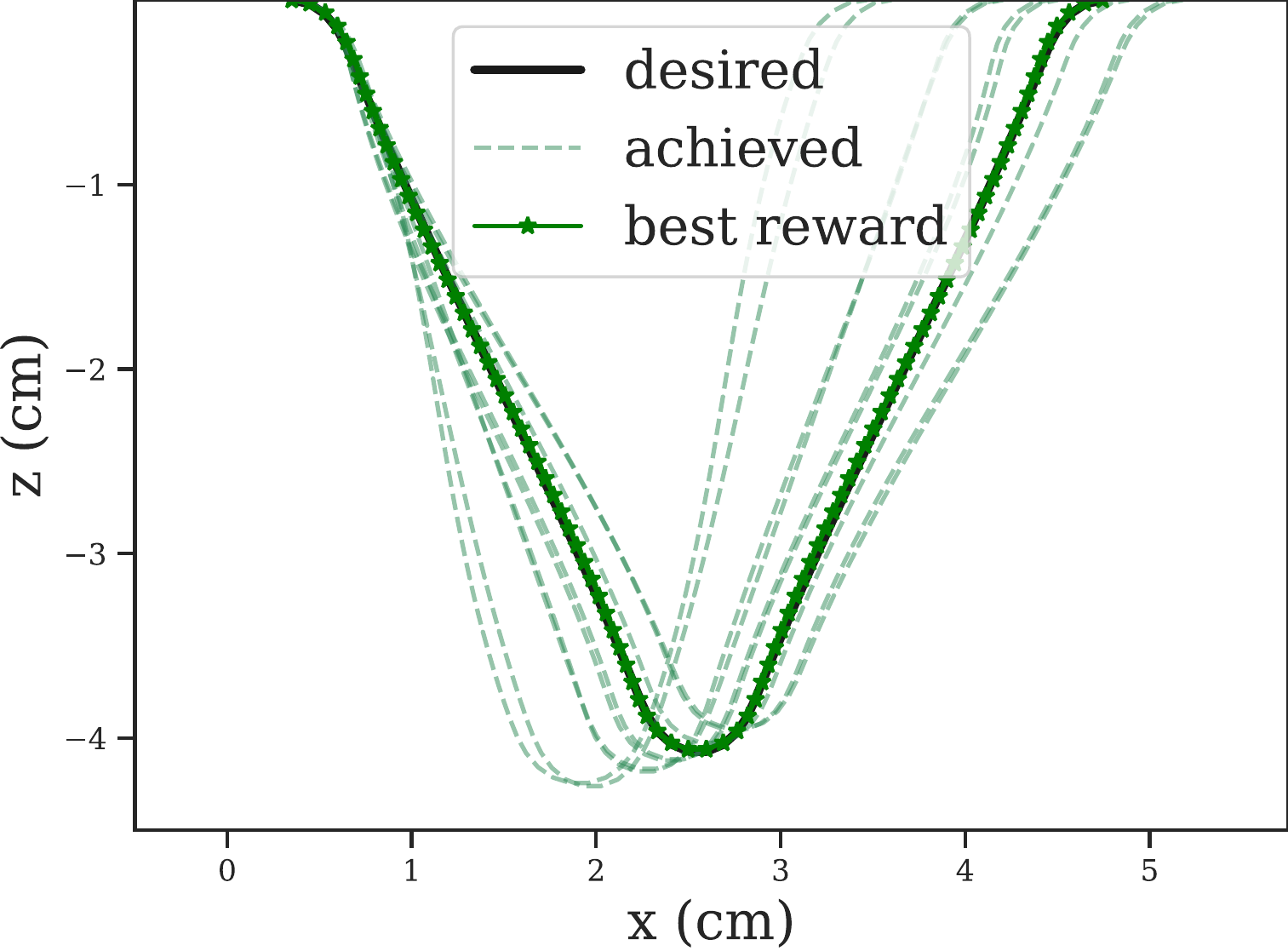}
    \end{subfigure}
    \begin{subfigure}{.49\columnwidth}
    \centering
    \includegraphics[width=\textwidth]{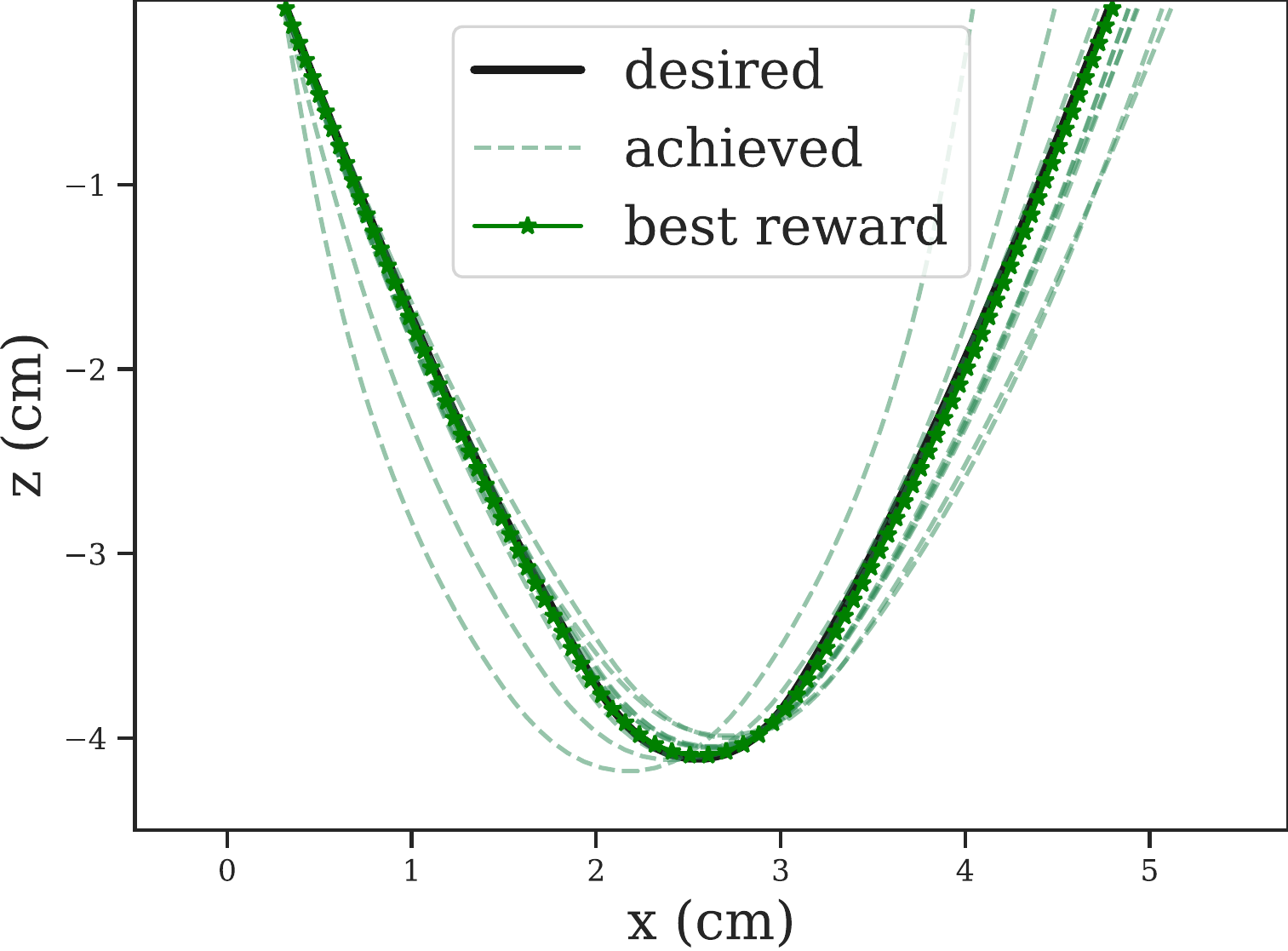}
    \caption{Hinge}
    \end{subfigure}
    \begin{subfigure}{.49\columnwidth}
    \centering
    \includegraphics[width=\textwidth]{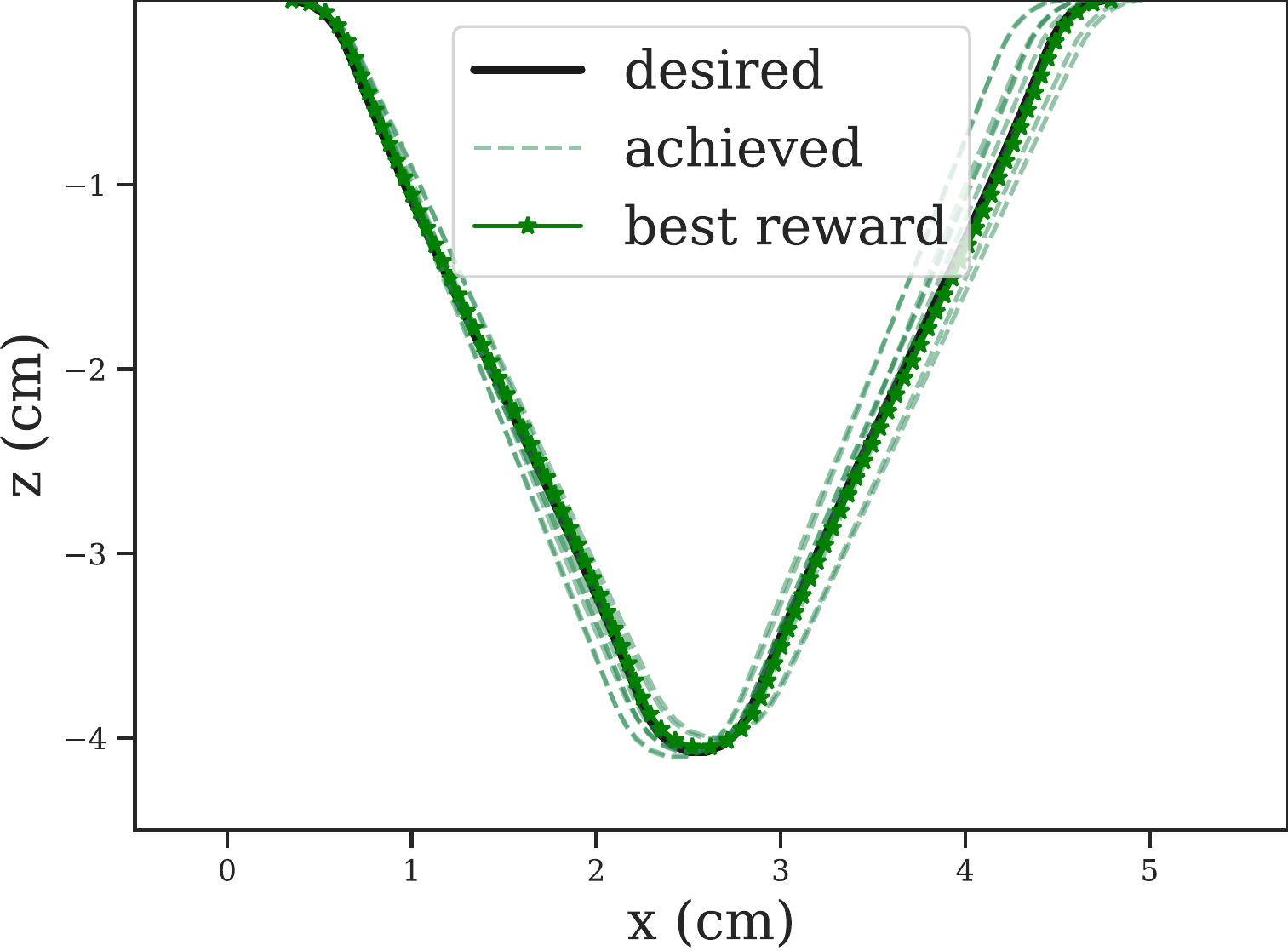}
    \caption{Lock}
    \end{subfigure}
    \caption{Final shapes achieved for 1 DoF problems. Three state representations and respective rewards are shown, namely reward \eqref{eq:position_reward} - top, reward \eqref{eq:curvature_reward} - middle and reward \eqref{eq:position_curvature_reward} - bottom. All ten trained policies are shown and shape leading to best reward is highlighted.}
    \label{fig:positions_v_curvature}
    \vspace{-0.5cm}
\end{figure}

Table \ref{tab:results} summarizes the results obtained with different reward functions, for an elastoplastic DLO. The Euclidean distance $L_T(\pmb{c}_T)$ was measured for each shape achieved after executing the learned control policies, with $T$ denoting the final time step. Reward \eqref{eq:position_reward} resulted in the best average distance however, as seen in Figure \ref{fig:positions_v_curvature} (top row), the learned policies do not reach the desired shapes with permanent deformation. Instead the DDPG agent gets stuck in a local reward maximum, where the gripper is in the correct position, but the DLO has an incorrect shape. On the other hand, using reward function \eqref{eq:curvature_reward} leads to the worst average distance between achieved and desired shapes. This is also evident in Figure \ref{fig:positions_v_curvature} (middle row) however, the policies successfully learn to reach the desired permanent deformation. Note that for the pinch grip, this reward function has another local maximum which is far from the goal shape. Finally, we weight the two previous functions in reward \eqref{eq:position_curvature_reward}. With $\alpha=0.5$, this results in an average distance similar to the one obtained with reward function \eqref{eq:position_reward}, while also reaching the correct plastic deformation, as shown in \ref{fig:positions_v_curvature} (bottom row). 

Finally, we evaluate the performance of DDPG with reward function \eqref{eq:position_curvature_reward} for 2 and 3 DoFs. Results show how the shape complexity affects the learning outcome. As seen in Figure \ref{fig:varying_dof}, the lock grip shape is more challenging to achieve. Note that for the 3D deformation, torsion $\pmb{\tau}$ was included in the state representation and reward function. As expected, the shapes are more difficult to reach, due to the increased state and action spaces, however the learned policies still learn to plastically deform the DLO.
\begin{figure}[t]
    \vspace*{0.2cm}
    \centering
    \begin{subfigure}{.45\columnwidth}
    \centering
    \includegraphics[width=\textwidth]{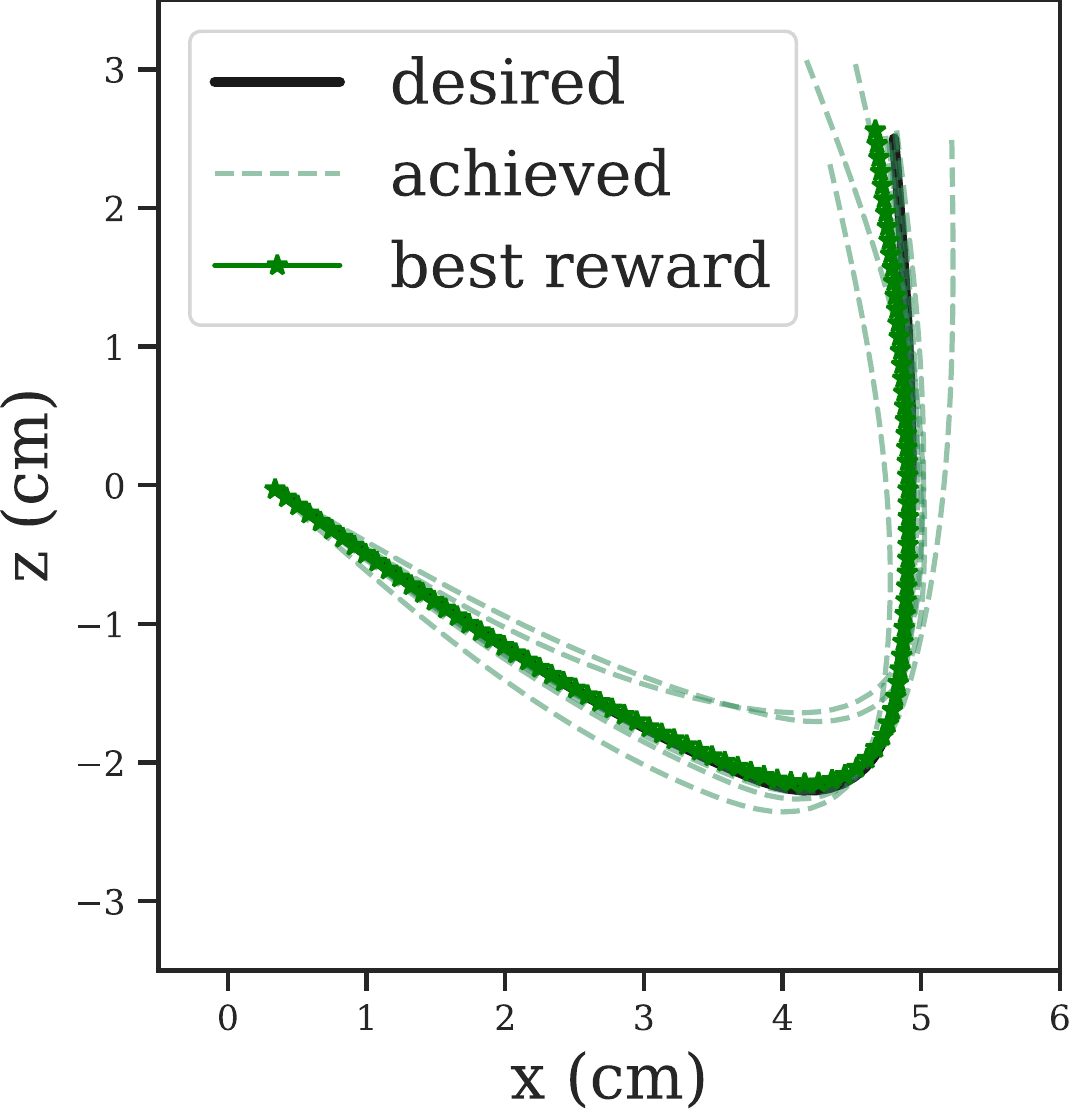}
    \end{subfigure}
    \begin{subfigure}{.45\columnwidth}
    \centering
    \includegraphics[width=\textwidth]{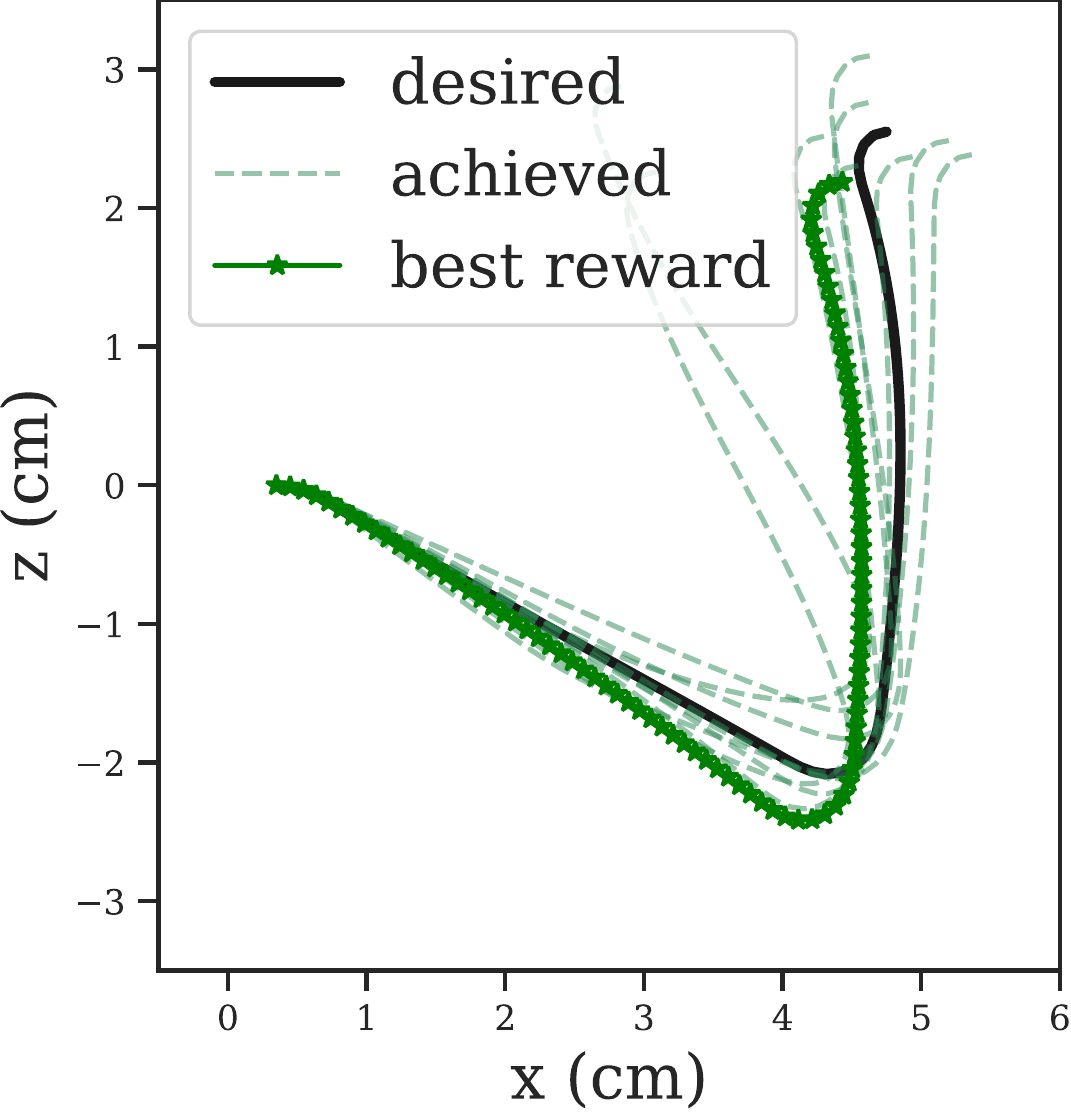}
    \end{subfigure}
    \begin{subfigure}{.45\columnwidth}
    \centering
    \includegraphics[width=\textwidth]{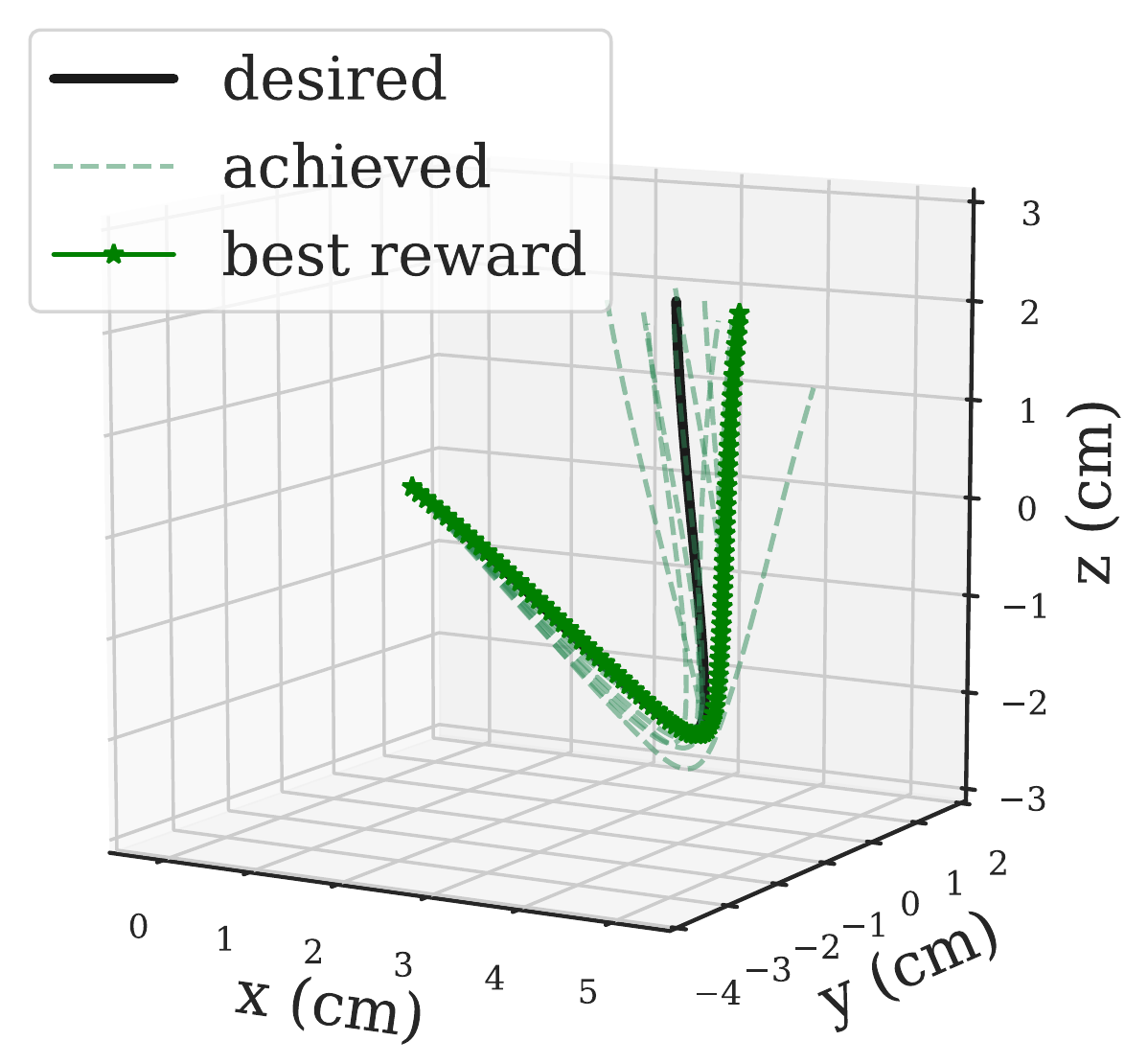}
    \caption{Hinge}
    \end{subfigure}
    \begin{subfigure}{.45\columnwidth}
    \centering
    \includegraphics[width=\textwidth]{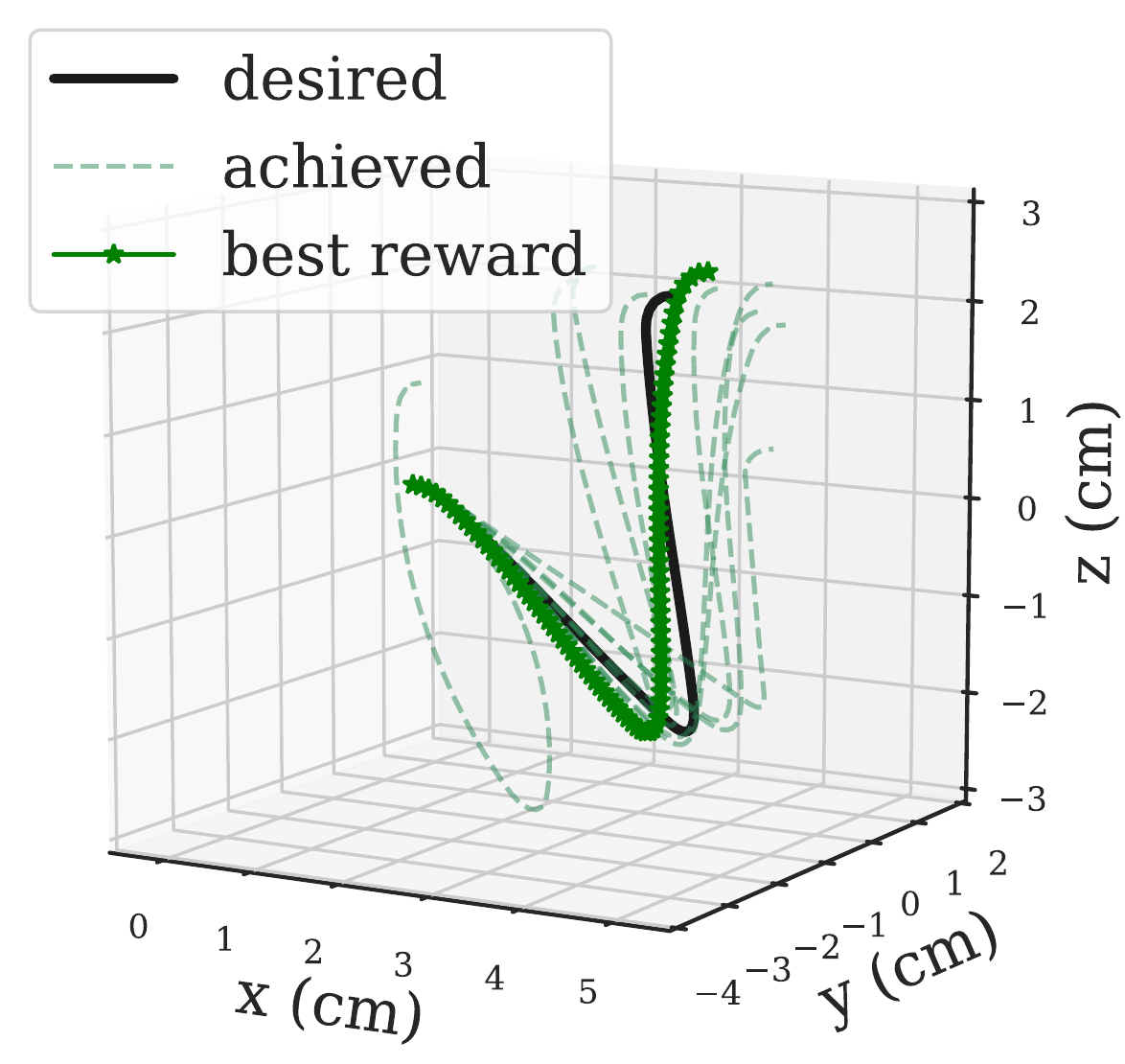}
    \caption{Lock}
    \end{subfigure}
    \caption{Results with higher DoFs. The best $ L_T (\pmb{c}_T)$ distances for the 2 DoF control (top) were: $[0.0030,0.0138]$; while for the 3 DoF control (bottom) they were: $[0.0183, 0.0228]$.}
    \label{fig:varying_dof}
    \vspace*{-0.5cm}
\end{figure}
\section{Concluding Remarks}
We have presented a new shape control problem for elastoplastic DLOs, highlighting its implementation challenges in the context of RL. We first presented the difficulties of designing a reward that encodes the correct goal, while also being resistant to local maxima. To that end we introduced a DLO shape representation, based on curvature and torsion of a discrete curve. This led to three alternative dense reward functions which were empirically compared.

Our results showed that for an elastic DLO, the 1 DoF shape control problem can be easily solved based on the Euclidean distance of the desired and achieved curves. However, with elastoplastic DLOs, the reward must also include a distance measure to the desired curvature and torsion. Finally, we evaluated the proposed weighted reward function with 2 and 3 controlled DoFs, leading to more challenging exploration, but still converging to reasonable shapes. 

For future work, we will use the proposed reward in multi-goal RL, to obtain a general solution to this type of shape control problems. Further, the problem of re-grasping and more complex shapes will be addressed. Ultimately the greatest challenge ahead is to bring our results from simulation into the real-world. This is planned to be done in a sim-to-real approach.


\section*{ACKNOWLEDGMENTS}
This work was supported by the Wallenberg AI, Autonomous Systems and Software Program (WASP) funded by the Knut and Alice Wallenberg Foundation. The simulations were developed with the support from Algoryx AB. The computations were enabled by resources provided by the Swedish National Infrastructure for Computing (SNIC) at C3SE partially funded by the Swedish Research Council through grant agreement no. 2018-05973.

\addtolength{\textheight}{-6.5cm}   

\bibliographystyle{ieeetr}
\bibliography{references.bib}

\end{document}